\documentclass[lettersize,journal]{IEEEtran}

\ifCLASSOPTIONcompsoc
  \usepackage[nocompress]{cite}
\else
  \usepackage{cite}
\fi

\usepackage[numbers]{natbib}

\usepackage{url}
\usepackage{CJKutf8}
\usepackage{algorithm}
\usepackage{subfigure}
\usepackage{xcolor}
\usepackage{booktabs} 
\usepackage{mathtools}
\usepackage{amsmath,bm}
\usepackage{makecell}
\usepackage{multirow}
\usepackage{multicol}
\usepackage{array}
\usepackage{comment}
\usepackage{amsfonts,amssymb}
\usepackage{textcomp}
\usepackage{fmtcount}
\usepackage{flushend}
\usepackage[skip=3pt]{caption}

\usepackage[noend]{algpseudocode}
\usepackage{algorithmicx,algorithm}

\usepackage{amsmath}
\usepackage{graphicx}
\usepackage{multirow}
\usepackage{multicol}
\usepackage{caption}
\usepackage{subfigure}
\usepackage{enumitem}
\usepackage{float}
\usepackage{makecell}
\usepackage{tabu}

\usepackage[T1]{fontenc}
\usepackage[utf8]{inputenc}
\usepackage{longtable}
\usepackage{color}
\usepackage{balance} 

\definecolor{mygray}{gray}{0.6}

\usepackage{tcolorbox}
\usepackage{cleveref}
\usepackage{bbm}
\usepackage{wrapfig}

\newcommand{\dtrain}{\mathcal{D}_{\text{train}}}

\newcommand{\seedset}{\mathcal{S}_{\text{seed}}}

\newcommand{\ours}{{\textsc{RetroPrompt}}}

\newcommand{\knn}{$k$NN}

\definecolor{right}{RGB}{0,128,96}
\definecolor{wrong}{RGB}{192,0,32}

\definecolor{blue}{rgb}{0,0,255} 

\ifCLASSOPTIONcompsoc
  \usepackage[nocompress]{cite}
\else
  \usepackage{cite}
\fi
\ifCLASSINFOpdf
\else
\fi
\begin{document}

\title{Retrieval-augmented Prompt Learning for  Pre-trained Foundation Models}

\author{
Xiang Chen, Yixin Ou, Quan Feng, Lei Li, Piji Li, Haibo Ye, Sheng-Jun Huang†, \\ Shuofei Qiao, Shumin Deng, Huajun Chen, Ningyu Zhang†

\IEEEcompsocitemizethanks{
\IEEEcompsocthanksitem{Xiang Chen, Piji Li, Haibo Ye and Sheng-Jun Huang are with MIIT Key Laboratory of Pattern Analysis and Machine Intelligence,
College of Computer Science and Technology, Nanjing University of Aeronautics and Astronautics.  E-mail: \{xiang\_chen, pjli, yhb, huangsj\}@nuaa.edu.cn.}
\IEEEcompsocthanksitem{Yixin Ou, Lei Li, Shuofei Qiao, Huajun Chen and Ningyu Zhang are with Zhejiang University. E-mail: \{ouyixin, leili21, shuofei, huajunsir, zhangningyu\}@zju.edu.cn.}
\IEEEcompsocthanksitem{Quan Feng is with Hunan Vanguard Group Corporation Limited. E-mail: jgxyfq@126.com.}
\IEEEcompsocthanksitem{Shumin Deng is with the National University of Singapore, Singapore.
E-mail: shumin@nus.edu.sg.}
\IEEEcompsocthanksitem{Corresponding author: Sheng-Jun Huang and Ningyu Zhang.}
\IEEEcompsocthanksitem{A previous version of this paper has been accepted as ``Decoupling Knowledge from Memorization: Retrieval-augmented Prompt Learning''~\cite{chen2022decoupling} in the 2022 Conference on Neural Information Processing Systems (NeurIPS 2022 Spotlight). 
This paper expands on the previous retrieval-augmented approach and applies it to computer vision tasks for multimodal learning.
In addition, we perform extensive experiments to confirm its effectiveness and generalizability across various benchmarks, supplemented by ablation studies and in-depth case studies.
The source code and datasets  can be accessed at \protect\url{https://github.com/zjunlp/PromptKG/tree/main/research/RetroPrompt}.}
}
}

\IEEEtitleabstractindextext{%

\begin{abstract}
The pre-trained foundation models (PFMs) have become essential for facilitating large-scale multimodal learning. Researchers have effectively employed the ``pre-train, prompt, and predict''  paradigm through prompt learning to induce improved few-shot performance.
However, prompt learning approaches for PFMs still follow a parametric learning paradigm. 
As such, the stability of generalization in memorization and rote learning can be compromised.
More specifically, conventional prompt learning might face difficulties in fully utilizing atypical instances and avoiding overfitting to shallow patterns with limited data during the process of fully-supervised training.
To overcome these constraints, we present our approach, named {\ours}, which aims to achieve a balance between memorization and generalization by decoupling knowledge from mere memorization. Unlike traditional prompting methods, {\ours} leverages a publicly accessible knowledge base generated from the training data and incorporates a retrieval mechanism throughout the input, training, and inference stages. This enables the model to actively retrieve relevant contextual information from the corpus, thereby enhancing the available cues.
We conduct comprehensive experiments on a variety of datasets across natural language processing and computer vision tasks to demonstrate the superior performance of our proposed approach, {\ours}, in both zero-shot and few-shot scenarios.
Through detailed analysis of memorization patterns, we observe that {\ours} effectively reduces the reliance on rote memorization, leading to enhanced generalization.

\end{abstract}
}




\maketitle

\IEEEdisplaynontitleabstractindextext

\IEEEpeerreviewmaketitle

\section{Introduction}

\IEEEPARstart{P}{re-trained} Foundation Models (PFMs) have achieved dramatic empirical success in various of domains such as natural language processing \cite{radfordimproving}, computer vision \cite{clip} and so on.
Notably, large-scale parametric foundation models 
have acquired a substantial volume of knowledge from multimodal sources, serving as fundamental infrastructure by demonstrating remarkable abilities with the ``pre-train, prompt, and predict'' paradigm \cite{DBLP:conf/iclr/000100LDC23}.
Prompt learning for PFMs has garnered growing research attention in recent years, based chiefly on few-shot data, for visual and language understanding.  

Typically, the ``prompt''  refers to a specific instruction or cue given to a machine learning model to guide it towards learning a specific task or to improve its performance on a specific task.
For instance, in the realm of natural language processing~\cite{gao2020making}, a prompt could be a sentence or phrase that provides context or specifies the type of output desired from the model; 
while in computer vision~\cite{clip} or multimodal learning~\cite{simvlm}, a prompt can guide the representation learning to combine information from various modalities including images and text, enabling models to enhance their learning efficiency by providing them with explicit guidance on what information to attend to.
To date, researchers have readily enjoyed themselves with the prompt learning for PFMs; evidence from emerging research has continuously proven its success in few-shot/zero-shot learning.
However, recent investigations \cite{DBLP:journals/corr/abs-2104-08786,DBLP:journals/corr/abs-2203-00902} have revealed that prompt learning with PFMs often exhibits unstable generalization in scenarios with extremely limited resources or emerging domains. This instability can be attributed, in part, to the inherent difficulty faced by parametric models in effectively \emph{learning rare or challenging patterns through rote memorization}, ultimately leading to suboptimal generalization performance.

\begin{figure}
	\centering
\includegraphics[width=0.45\textwidth]{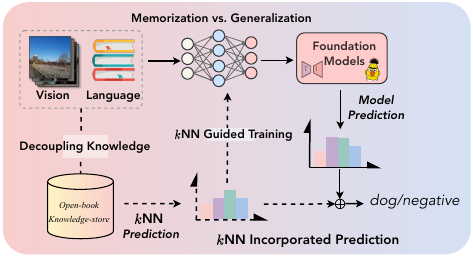}
\caption{
Decoupling knowledge from memorization.
To achieve a harmonious balance between generalization and memorization in prompt learning, we put forward a method that separates knowledge from mere memorization. Our approach involves creating a knowledge-store accessible for reference and retrieval throughout the training and inference phases.
} 
\label{fig:motivation}
\end{figure}

Prior work has established metaphors for conceptualizing the training-test procedures in prompt learning akin to {\it closed-book examination} and {\it page-by-page memorization}~\cite{meng2021gnnlm}. Specifically, conventional prompt learning faces challenges either rote memorizing atypical cases under full supervision or overfitting shallow patterns with limited data~\cite{elangovan-etal-2021-memorization}.
Recent research~\cite{feldman2020does} provides evidence supporting the long-tail theory, which suggests that training instance often follows a long-tailed distribution characterized by small sub-populations containing rare examples, PFMs may predict through memorizing these outliers rather than generalizing patterns - indicating a reliance on rote memorization over truly learning representations. This reliance on memorization contrasts with the objective of effectively utilizing knowledge from varied instances to achieve robust generalization. Addressing such limitations motivates the proposed approach of augmenting prompting with context retrieval to balance memorization and generalization.

Rote memorization's limitations encourage us to seek inspiration from the human learning process, especially the principle of \emph{`learning by analogy'}, and to acknowledge the wisdom in the saying, \emph{`The palest ink is better than the best memory'}.
Interestingly, humans demonstrate exceptional abilities in associative learning, harnessing profound memories to strengthen pertinent abilities,
which enables them to tackle tasks with few or no prior examples.
We aim to enhance prompt learning generalization by leveraging retrieval and association, inspired by observations of its limitations. 
Our primary viewpoint suggests that tackling these challenges can be effectively mitigated by dissociating knowledge from mere memorization through the utilization of an {\it open-book knowledge-store} derived from the training data.
By referencing relevant knowledge, we can provide the model with a robust signal for striking a balance between generalization and memorization, thereby significantly mitigating the challenges mentioned above.


Specifically, we propose a novel retrieval-augmented framework, called \textbf{\ours}, which builds upon prompt learning (Figure~\ref{fig:motivation}). 
To decouple knowledge from pure memorization, we introduce an open-book knowledge store $\left(\mathcal{K},\mathcal{V}\right)$, 
composed of \emph{key-value} pairs extracted from the training data, with \emph{keys} representing prompt-based example embeddings and \emph{values} corresponding to label words. 
To incorporate the retrieved knowledge into the model input,
\textbf{Firstly}, we employ a non-parametric algorithm \knn{} to determine the difficulty level of instances. 
This is accomplished by comparing the input query with the knowledge-store and introducing a scaling factor during training to amplify the influence of challenging instances identified through the \knn{} process.
\textbf{Furthermore}, the outcomes of the \knn{} approach are incorporated at the output of the prompt-based fine-tuning model (PFM) head, contributing to the masked prediction process. 
During the inference process, the model employs linear interpolation to combine the output obtained from prompt learning with the non-parametric nearest neighbor distribution. This integration involves leveraging cues from the Top-$k$ nearest reference instances within the $\left(\mathcal{K},\mathcal{V}\right)$ pair.
In the context of language understanding, we propose the incorporation of neural demonstrations, and the concatenation of this with input instances can occur at the embedding layer.
The objective of this approach is to enhance the performance to generalize across different tasks or scenarios.
\textbf{Besides}, the \knn{} results provide input to the prediction head after prompt tuning, influencing outputs through linear interpolation with the nearest reference instances in the knowledge base.
In the context of language understanding tasks, we also introduce neural demonstrations concatenated at the embedding layer to augment inputs and enhance generalization capabilities. 
For visual understanding, we adopt the architecture of CLIP \cite{clip}, but solely train prompts by creating a query-key knowledge-store from few-shot supervisions to obtain prompt weights.

We evaluate {\ours} on various datasets, including language and visual understanding.
The significant performance improvements observed in both language and visual understanding tasks in scenarios involving zero-shot and few-shot learning settings affirm the efficacy of our systemic retrieval mechanism in enhancing model generalization with limited data. Additionally, our {\ours} approach demonstrates robustness in handling atypical instances within the fully-supervised setting, particularly in scenarios characterized by the long-tail distribution.
To gain further insights into the memorization process, we utilize self-influence \cite{koh2017understanding} as our scoring function for analyzing memorization across fine-tuning, prompt learning, and our {\ours} approach. Comprehensive examination reveals the following key findings: 1) training samples with the highest memorization scores predominantly comprise atypical instances; 2) by decoupling knowledge from memorization and mitigating the rote memorization tendencies of pre-trained foundation models (PFMs), our proposed {\ours} approach surpasses both fine-tuning and conventional prompt-tuning methods in terms of performance.
In summary, our approach presents a promising direction for enhancing the generalization capabilities of prompting PFMs through the decoupling of knowledge from mere memorization, which opens up new avenues for future research. Overall, our work delivers several noteworthy contributions to the field:

\begin{itemize}
\item  We propose {\ours} as a novel approach for language and visual understanding tasks. Our motivation is to decouple knowledge from pure memorization, enabling the model to find a balance between generalization and memorization.
\item The proposed {\ours} incorporates a retrieval augmentation mechanism throughout the input, training, and inference stages. It leverages pertinent contexts extracted from the training corpus as informative cues to optimize PFM performance.
\item We conduct comprehensive experiments various datasets related to visual and language understanding tasks,
showcasing the impressive few-shot/zero-shot capabilities and generalization of {\ours}.
\end{itemize}

\section{Related Work}

\begin{figure*}       
    \centering
    \includegraphics[width=0.85 \textwidth]{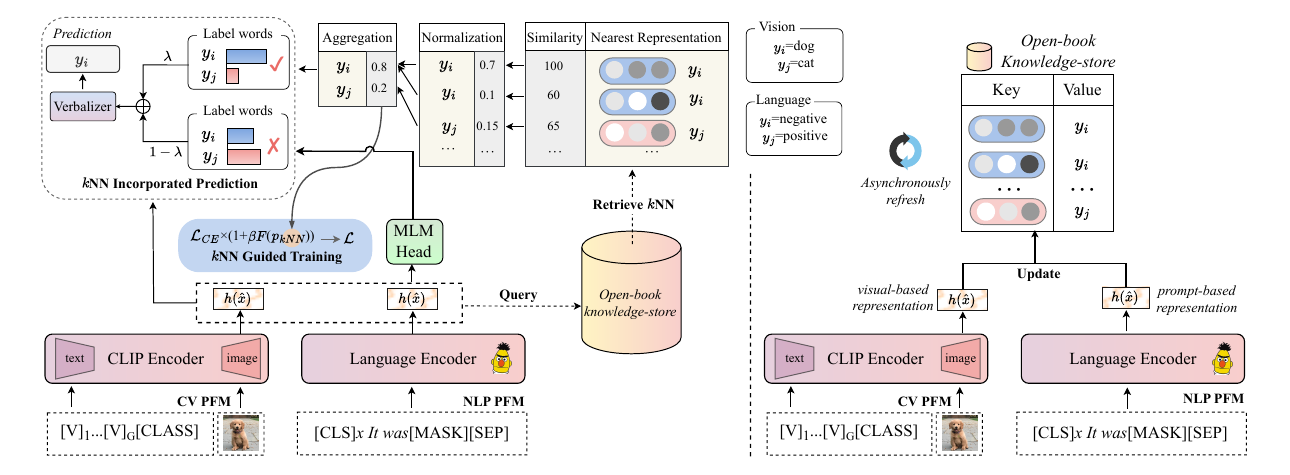}
\caption{
Illustration of {\ours}. 
}
\label{fig:arc}
\end{figure*}

\subsection{Pre-trained Foundation Models}





In the present era characterized by the abundance of big data, pre-trained foundation models (PFMs) ~\cite{DBLP:journals/taslp/QinXWZC23} play a critical role in the domain of AI.
By employing the pre-training methodology, PFMs undergo training with vast amounts of data and tasks, enabling seamless fine-tuning for diverse downstream applications.
Originally stemming from transfer learning in Computer Vision (CV) tasks, pre-training techniques have been recognized for their effectiveness, such as ViT~\cite{vit}.
Previous research in the NLP) domain has demonstrated that pre-trained language models such as  GPT-3~\cite{DBLP:conf/nips/BrownMRSKDNSSAA20}, LLaMA \cite{DBLP:journals/corr/abs-2302-13971} and ChatGPT possess the capacity to capture comprehensive hierarchical knowledge supporting tasks like long-range dependencies.
Joint vision-language learning has also proven effective, such as CLIP~\cite{clip} showcasing zero-shot classification abilities through multi-modal pre-training.

\subsection{Prompting PFMs}

The advent of GPT-3 has paved the way for the advancement of prompt learning techniques~\cite{DBLP:journals/emnlp/qinlibo}, effectively bridging the gap between the masked language modeling objective employed by PFMs and the downstream fine-tuning objectives.
In the field of NLP, prompt learning has demonstrated remarkable performance across various tasks, including text classification~\cite{schick2020automatically} and information extraction\cite{chen21knowprompt}, particularly in the few-shot setting.
Prompt learning techniques have been instrumental in advancing cross-modal tasks and visual understanding within the domain of computer vision, such as image classification \cite{clip}, visual grounding \cite{cpt,CoCoOp} and image captioning \cite{DBLP:conf/acl/Jin0SC022}, among others.
Furthermore, continuous prompts~\cite{p_tuning} have been introduced to mitigate prompt engineering, appending a sequence of trainable embeddings as prompts to the input.

It is important to note that our work diverges from previous prompt learning approaches, which primarily focus on optimizing prompts. 
Instead, our study focuses on the systematic retrieval of relevant examples from the training data as a means to improve the effectiveness of prompt learning.

\subsection{Retrieval-enhanced PFMs}

Retrieval-enhanced approaches\cite{rag_survey} have been applied to various domains, including NLP, CV and multimodal learning.
Some recent studies~\cite{gao2020making,DBLP:journals/corr/abs-2101-06804} adopt the approach of retrieving a small number of training examples and presenting them as discrete demonstrations within a natural language prompt for PFMs, which can induce informative prompts thus making better few-shot performance.
Another research area of retrieval augmentation~\cite{DBLP:conf/nips/LewisPPPKGKLYR020,DBLP:journals/corr/abs-2112-08633,RETRO} focuses on the retrieval of valuable clues from external knowledge corpora, such as Wikipedia, to support specific tasks like open-domain question answering.
Besides, semi-parametric methods~\cite{DBLP:conf/iclr/KhandelwalLJZL20,he2021efficient,DBLP:conf/iclr/KhandelwalFJZL21,alon2022neurosymbolic} have emerged employing the classic non-parametric $k$-nearest neighbors classifier, relying on representation similarity, to enhance the performance of PLMs across diverse tasks.
By comparison, our proposed {\ours} has two main differences from the above-mentioned works: (1) our objective is not limited to the enhancement of inference alone, we instead strive to develop a holistic retrieval mechanism that operates throughout the stages of input, training, and inference.
(2) we focus on contributing a general solution for both language and visual understanding tasks, thus, pluggable to various previous prompt learning approaches for different foundation models.



\section{Preliminaries}
\label{background}


\subsection{Prompt Learning for PFMs}

Prompt learning has gained significant attention since the emergence of GPT-3~\cite{DBLP:conf/nips/BrownMRSKDNSSAA20}. 
A series of research works~\cite{schick2020exploiting, schick2020automatically,shin2020eliciting} have emerged, indicating that prompt learning demonstrates a more effective utilization of knowledge within PFMs compared to traditional fine-tuning methods.
Let $\mathcal{M}$ represent the PFM and $\mathcal{T}$ denote the template function utilized in prompt learning. The language and visual understanding can be described as follows.

\hspace*{\fill} \\
\noindent
\textbf{Prompting Language Understanding.}
In the task of text classification, we are provided with an input query sentence $\bm{x} = (x_0,x_1,...,x_n)$,  and our objective is to assign it the label ${y} \in \mathcal{Y}$.
In order to convert the given task into a masked language modeling (MLM) problem that incorporates \textit{cloze-style} objectives, a template function $\mathcal{T}$ is utilized to insert text fragments into the input sequence $\bm{x}$, resulting in $\hat{\bm{x}} = \mathcal{T}(\bm{x})$.
$\hat{\bm{x}}$ is the representation of the input to $\mathcal{M}$ that incorporates a {\tt[MASK]} token.
For example, given a task to categorize the text string $\bm{x}$ = ``This food is quite unpalatable.'' under the labels \textsc{Positive} (assigned the label 1) or \textsc{Negative} (assigned the label 0), we would encapsulate it as follows:
\begin{equation}
\hat{\bm{x}}=
\texttt{[CLS]} \bm{x} \ \text{It was \texttt{[MASK]}} \texttt{[SEP]}
\end{equation}
Furthermore, we establish a verbalizer function, denoted as $f\colon \mathcal{Y} \mapsto \mathcal{V}$, that links the label space $\mathcal{Y}$ to words within the vocabulary, thereby constructing the set of \emph{label words} $\mathcal{V}$.
The primary module of $\mathcal{M}$ generates a sequential representation of $\hat{\bm{x}}$, and we extract the hidden representations corresponding to the position of the \texttt{[MASK]} token as the contextual representation $\bm{h}_{\hat{\bm{x}}} \in \mathbb{R}^d$, with $d$ denoting the dimension of hidden states.
The masked language modeling (MLM) head of $\mathcal{M}$ processes $\bm{h}{\hat{\bm{x}}}$ to compute the probability $P\mathcal{M}(\texttt{[MASK]}=v|\hat{\bm{x}})$ for each word $v$ in the vocabulary, representing the likelihood of inserting that word at the \texttt{[MASK]} position.
We denote $\mathcal{V}y$ as the subset of $\mathcal{V}$ that corresponds to a particular label $y$, satisfying the condition:
$\cup{y\in\mathcal{Y}} \mathcal{V}_y = \mathcal{V}$.
Ultimately, the computation of the probability distribution for the label $y$ is performed as follows:
\begin{equation}
\begin{aligned}
P(y|\bm{x}) = g\left(P\mathcal{M}(\texttt{[MASK]}=v|\mathcal{T}(\bm{x})), v\in\mathcal{V}_y\right) ,
\end{aligned}
\label{eq:pmscore}
\end{equation}
where the function $g$ represents the conversion of label word probabilities into class probabilities.

\hspace*{\fill} \\
\noindent
\textbf{Prompting Visual Understanding.} \quad 
We adopt prompt learning based on pre-trained CLIP for image classification, which  takes a query image $\bm{x}$ as input and assigns it to the label ${y} \in \mathcal{Y}$. 
In contrast to the pre-trained language model that transforms the language understanding task into a MLM issue, the alignment of image and text embedding spaces is achieved through the utilization of a contrastive loss, aiming to  optimize the cosine similarity to be at its peak for paired sets recognized as matches, while minimizing it for all unpaired sets in an image-text batch.
Regarding CLIP, which encompasses the visual encoder $\mathcal{M}_v$  and the text encoder $\mathcal{M}_t$, the function $\mathcal{T}$ integrates prompt elements alongside the relevant class token into the text encoder $\mathcal{M}_t$ as follows:

\begin{equation}
 \boldsymbol{t}=[\mathrm{V}]_1[\mathrm{~V}]_2 \ldots[\mathrm{V}]_G[\mathrm{CLASS}].
\end{equation}
The vector  $[\mathrm{V}]_g$, where $g$ ranges from 1 to the hyperparameter  $G$, aligns dimensionally with word embeddings (for instance, 512 in CLIP). Here,  $G$ delineates the count of context tokens.
It is notable this context is shared across classes, referred to as the unified context, differing from class-specific contexts. 

For an image $\boldsymbol{x}$, let $\bm{h}_{\hat{\bm{x}}}$ represent the image features extracted with the image encoder, and let the image features derived using the image encoder be denoted by $\bm{h}_{\hat{\bm{x}}}$, and consider $ \{\boldsymbol{w}_i\}_{i=1}^{G+1}$ as the collection of weight vectors produced via the text encoder.
Represent the quantity \( K \) as the count of categories. Each \( \boldsymbol{w}_i \) is derived from a prompt potentially adopting the aforementioned structure, substituting the class descriptor with distinct terms like ``dog'', ``cat'', or ``car''. To generate a classification weight vector symbolic of a visual notion, we input \( \hat{\bm{x}} \) into the text encoder \( \mathcal{M}_t \). Subsequently, we compute the predictive probability with distance function as:

\begin{equation}
P(y=i \mid \boldsymbol{x})=\frac{\exp \left(<\mathcal{M}_t\left(\boldsymbol{t}_i\right), \bm{h}_{\hat{\bm{x}}}>/ \tau\right)}{\sum_{j=1}^K \exp \left(<\mathcal{M}_t\left(\boldsymbol{t}_j\right), \bm{h}_{\hat{\bm{x}}}>/ \tau\right)}
\end{equation}
For every given prompt \( \boldsymbol{t}_i \), the class identifier is substituted by the word embedding vector(s) corresponding to the \( i \)-th class label.
Training minimizes the standard cross-entropy classification loss, allowing gradients to backpropagate through the text encoder $\mathcal{M}_t$ and fully leverage its learned knowledge to optimize the context. Continuous representations explore the embedding space thoroughly, enhancing the learning of relevant contexts.

\subsection{Motivation: Essential Issues of Prompting for PFMs}

Recent research studies \cite{p_tuning, DBLP:journals/corr/abs-2104-08786, DBLP:journals/corr/abs-2203-00902} have found that prompt learning using pre-trained models often fails to generalize well in extremely low-resource settings or emerging domains. 
A plausible rationale might be that parametric models encounter challenges in effectively assimilating infrequent or intricate patterns via direct memorization, leading to less-than-ideal generalization capabilities.
To gain deeper insights into this matter, the training procedure can be equated to studying from a book, while the testing stage resembles taking a test. The traditional method of prompt learning, involving training with batch datasets, bears resemblance to ``sequential page memorization" and facing a ``non-referenced test''.
In this context, vanilla prompt learning encounters difficulties in retaining uncommon examples in a fully-supervised environment or succumbing to superficial patterns when dealing with sparse data \cite{memory}.
These observations inspire us to explore leveraging retrieval to enhance prompt learning through the aspect of decoupling knowledge from memorization.
We argue that not all ``modeledge'' can be efficiently learned in PFMs; thus, retrieval augmentation can strike a balance between generalization and memorization, also bringing opportunities to update/add new knowledge in the updating era.

\section{The Proposed Approach: {\ours}}
We present {\ours}, a novel method that builds upon the dense retriever and integrates an open-book knowledge-store to distinguish between knowledge absorption and sheer memorization.
Illustrated in Figure \ref{fig:arc}, {\ours} comprises three key parts: a dense retriever (\textit{cf. }Section \ref{sec:store}), \knn{} guided training (\textit{cf. }Section \ref{sec:knn-train}), \knn{}-based probability estimation for \textit{cloze-style} prediction (\textit{cf. }Section \ref{sec:knn-test}), and can be adapted to both language and visual understanding (\textit{cf. }Section \ref{sec:task-adaptation}).

\subsection{Dense Retriever}
\label{sec:store}
\subsubsection{Open-book Knowledge-store}
In the initial phase of our suggested framework, we establish a knowledge-store designed to aid retrieval, disentangle direct memorization, and encapsulate the semantic core of samples from training set \( \mathcal{C} \).
To achieve this, we utilize an encoder to generate embeddings of the samples in $\mathcal{C}$, which are then utilized for constructing the knowledge-store. 
For each training example $\left(\bm{c}_i,{y}_i\right) \in \mathcal{C}$, we create a key-value pair $(\bm{h}_{\hat{\bm{c}}_i},v_i)$, where $\hat{\bm{c}}_i = \mathcal{T}({\bm{c}_i})$ and $\bm{h}_{\hat{\bm{c}}_i} \in \mathbb{R}^d$ represents the output embedding of the PFM, and $v_i = f(y_i)$ denotes the corresponding label word.
For the training data \( \mathcal{C} \) and its \( i \)-th instance \( \left(\bm{c}_i,{y}_i\right) \), a key-value tuple \( (\bm{h}_{\hat{\bm{c}}_i},v_i) \) is formed. Here, \( \hat{\bm{c}}_i=\mathcal{T}({\bm{c}_i}) \), the element \( \bm{h}_{\hat{\bm{c}}_i} \in \mathbb{R}^d \) signifies the resultant embedding from the pre-trained foundation model (PFM), and \( v_i=f(y_i) \) stands for the designated label word of the \( i \)-th sample.
Importantly, our knowledge-store deviates from the method outlined in kNN-LM~\cite{DBLP:conf/iclr/KhandelwalLJZL20}, which builds upon a shifting generative corpus and its respective tokens.
Instead, the knowledge-store is tailored for prompt learning purposes. Every tuple of $(\bm{h}_{\hat{\bm{c}}},v)$ is accommodated within a key-value database represented by $\left(\mathcal{K},\mathcal{V}\right)$. Here, $\bm{h}_{\hat{\bm{c}}}$ acts as the designated \emph{key}, and $v$ assumes the role of the corresponding \emph{value}. The establishment procedure is articulated as:

\begin{equation}
\begin{aligned}
\left(\mathcal{K},\mathcal{V}\right) =
{
\left(\bm{h}_{\hat{\bm{c}}_i},v_i\right) \mid \left({\bm{c}_i},y_i\right)\in \mathcal{C}
}
\label{eq:kv}
\end{aligned}
\end{equation}

In Equation~\ref{eq:kv}, the `|' operator denotes the `such that' condition, which is used to define a set comprehension.
The PFM serving as the encoder is not frozen but is actively fine-tuned during training. As the model's parameters are updated, the embeddings stored as keys become outdated, as they no longer reflect the current state of the encoder. To address this, the key-value pairs are updated.
This knowledge-store is amenable to dynamic alterations, encompassing edits, augmentations, or eliminations of samples. 
It's imperative to note that for few-shot scenarios, the knowledge-store is assembled exclusively from the few-shot training datasets, rather than a comprehensive training corpus.

\subsubsection{Efficient Searching}
To ensure efficient retrieval in the presence of potentially large training data $\mathcal{C}$, we employ strategies for optimized search operations.
Subsequent to the development of the previously mentioned open-book knowledge repository, we formulate the matrix \( \mathbf{D} \in \mathbb{R}^{|\mathcal{C}|\times d} \) to serve as a reference for the training samples.
Upon receiving a query set \( Q \), our first step involves encoding each query instance via the template transformation function \( \mathcal{T}(\cdot) \). This results in an assembly of prompt-oriented query vectors \( \bm{h}_{\hat{q}} \) intended for on-the-fly retrieval enhancement.
Following that, we utilize the aforementioned query vectors to identify the most analogous instances in the index \( \mathbf{D} \) by leveraging the maximum inner product search (MIPS) method.
During the retrieval stage, we turn to FAISS~\cite{DBLP:journals/tbd/JohnsonDJ21}, an adept open-source library tailored explicitly for rapid nearest neighbor search operations.

\subsubsection{Concurrent Refresh of the Knowledge-store}
\label{refresh}
Considering that the contextual representation of instances may vary due to the continual parameter updates in the PFM during neural demonstration, the search index associated with the demonstration could become outdated and go ``stale'' after the gradient update. 
To mitigate this limitation, we suggest a periodic ``refreshing'' of the retrieval index by asynchronously re-indexing and re-embedding all embeddings within the open-source knowledge-store after every  $j$ training cycles~\footnote{In our experiments, we refresh for each epoch.}.  In \textit{cf. }Section~\ref{ablation}, we empirically demonstrate the performance improvement resulting from this procedure.

\subsection{Utilizing \knn{} to Steer Training}
\label{sec:knn-train}
Eager learning models, like PFMs, aim to establish a universal function that connects text with a semantical label domain.
On the other hand, lazy learners like $k$-nearest neighbor classifiers aim to approximate the neighborhoods around test examples. Leveraging the classification results of \knn{} as \textbf{prior external knowledge} to guide the parameter adaptation of PFMs during training (referred to as \knn{}-train) is an intuitive approach. 
This approach is particularly effective for addressing challenging examples, which often correspond to atypical samples. In our method, we distinguish between simple and hard instances based on \knn{} predictions. 
For the query instance denoted as $\bm{q}_t$ at time $t$, the corresponding query vector $\bm{h}{q_t}$ is employed to extract the top $k$ closest instances, represented by $\mathcal{N}$, from the open-book knowledge-store $\left(\mathcal{K},\mathcal{V}\right)$. This extraction is based on the similarity metric $d(\cdot, \cdot)$.
Commonly, the function $d(\cdot, \cdot)$ uses the inner product as its similarity measure.
Subsequently, we determine the distribution across the neighboring data points by implementing the softmax function on their associated similarities. The cumulative probability for each label word is then calculated based on its frequency in the fetched targets, as illustrated in Equation~\ref{eq:knnscore}:

\begin{equation}
\small
\begin{aligned}
 P_{\text{\knn{}}}\left(y \mid \bm{q}_t \right)  & \propto
 \sum_{\left(\bm{c}_i,y_i\right)\in \mathcal{N}} \mathbbm{1}_{y=y_i} \exp\left(d\left( \bm{h}_{\hat{\bm{q}}_t}, \bm{h}_{\hat{\bm{c}}_i}\right)\right).
\label{eq:knnscore}
\end{aligned}
\end{equation}

Considering the probability $p_{k\text{NN}}$ associated with the prediction of the query instance $\bm{q}_t$ as the \textbf{reference class} (indicating the likelihood of the reference class in $P_{k\text{NN}}$), our approach integrates \knn{} to steer the prompt learning mechanism. The \knn{} navigator modifies the respective loss for instances rightly classified or misclassified as detected by \knn{} by altering the weighting of the cross-entropy loss $\mathcal{L}{CE}$. The negative log-likelihood serves as the adjusting element $F(p_{k\text{NN}})$. The resultant loss, represented as $\mathcal{L}$, is outlined as:
\begin{equation}
\label{eq:joint}
\small
F(p_{k\text{NN}}) = - \log{}(p_{k\text{NN}}), \quad
\mathcal{L} = \left(1 + \beta F(p_{k\text{NN}}) \right)\mathcal{L}_{CE}.
\end{equation}
In this context, $\beta$ acts as a scalar defining the weightage of each loss component. Crucially, $p_{k\text{NN}}$ is determined via the \emph{leave-one-out} distribution across the training dataset, ensuring that individual training samples do not reference themselves. The underlying principle for the adjustment element draws from Focal-loss~\cite{focal_loss}, though our primary intent is to harness the \knn{} outcomes to bolster language model training.

\subsection{\knn{}-Based Probability for \textit{Cloze-Style} Inference}
\label{sec:knn-test}
Beyond the \knn{}-oriented training methodology (termed as \knn{}-train), we incorporate a \knn{}-driven probability mechanism tailored for \textit{Cloze-style} prediction in the inference stage.
This approach enables PFMs to retrieve nearest neighbors for decision-making, rather than relying solely on memorized parameters for predictions.
Considering the non-parametric distribution $P_{k\text{NN}}$ of the query instance $\bm{q}_t$ projected to be $y$, we recalibrate $P(y\mid \bm{q}_t)$ by melding $P_{k\text{NN}}$ with forecasts from the pre-trained foundational model (PFM), symbolized as $P_\mathcal{M}$, influenced by a factor $\lambda$. This combined approach yields the ultimate probability associated with the label:

\begin{equation}
\begin{aligned}
\label{eq:lambda}
P(y \mid \bm{q}_t)= (1-\lambda)  P_\mathcal{M}(  y|\mathcal{T}(\bm{q}_t)) + \lambda P_{k\mathrm{NN}}(y \mid \bm{q}_t). 
\end{aligned}
\end{equation}
Diverging from methods like $k$NN-LM~\cite{DBLP:conf/iclr/KhandelwalLJZL20,he2021efficient} that chiefly use token retrieval to bolster language models, we concentrate on harnessing the \knn{} distribution anchored in prompts as a guiding reference during inference. This approach allows the prediction procedure to operate akin to an {\it open-book} assessment in the realm of prompt learning.

\subsection{Customization for Specific Tasks}
\label{sec:task-adaptation}
\subsubsection{Language Understanding}
For tasks that require understanding language, such as evaluating sentiments or relation extraction, we use a method we call {\ours}. This method is essentially a masked token prediction technique. To facilitate analogy-based learning within the PFMs using the knowledge-store, we incorporate neural demonstrations.
We append these demonstrations to the input instance during the embedding phase, thereby augmenting the broad generalization potential of our proposed approach. For a given query instance, represented as $\bm{q}_t$, the first step involves leveraging the prompt-based representation, $\bm{h}_{\hat{q}_t}$, to probe cached representations present within the knowledge-store.
Following this, we extract $m$ closest neighbors, represented as $\{ \{\bm{c}^{(1)}_{1}, ..., \bm{c}^{(1)}_{m}\}, ..., \{\bm{c}^{(L)}_{1}, ..., \bm{c}^{(L)}_{m}\}\}$, for every class. In this context, the superscript $L$ denotes the overall class count, and $\bm{c}_{i}^{(l)}$ refers to the $i$-th closest neighbor of the $l$-th class.
After identifying the top-$m$ candidates for each class, we incorporate the corresponding representations $\bm{h}_{\bm{\hat{c}}{i}}^{(l)}$, along with the label word $v^{(l)}$ retrieved from the knowledge store, into the encoding process for demonstrative learning.
As $\bm{h}_{\bm{\hat{c}}_{i}}^{(l)}$ already exists in vector form, it's judicious to coalesce the $m$ neighboring vectors of each class based on similarity. 
This demonstrative data is then melded with the input $\hat{\bm{x}}$ immediately after the word embedding stage of $\mathcal{M}$ as:

\begin{equation}
\small
\begin{aligned}
    \mathcal{I} &= {e}(\hat{\bm{x}}) \oplus 
    [\sum_{i \in [1:m]}\alpha_{i}^{(1)} \bm{h}_{\hat{\bm{c}}_i}^{(1)},
    {e}(v^{(1)})] \\
    & \oplus  ... \oplus 
    [\sum_{i \in [1:m]}\alpha_{i}^{(L)} \bm{h}_{\hat{\bm{c}}_i}^{(L)},
    {e}(v^{(L)})] ; 
    \\
    \alpha_i^{(l)} &= \frac{e^{
    \bm{h}_{\hat{\bm{q}}}
    \cdot \bm{h}_{\hat{\bm{c}}_i}^{(l)}}}
    {\sum_{i \in [1:m]} e^{\bm{h}_{\hat{\bm{q}}}
    \cdot \bm{h}_{\hat{\bm{c}}_i}^{(l)}}}.
\end{aligned}
\end{equation}
In the above equation, 
the function ${e}(\cdot)$ denotes the word embedding layer within the model $\mathcal{M}$.
The operator $\oplus$ signifies the fusion of input sequences. The coefficient ${\alpha}_{i}^{(l)}$ stands for the softmax score of the $i$-th retrieval linked with the label of the $l$-th class, reflecting its significance to $\hat{\bm{q}}$.
$\mathcal{I}$ represents the sequence features input into the next layer of the prompt-based fine-tuning model. 
The equation illustrates that the demonstration representation is encoded through a weighted aggregation of the retrieved representations.
This allows the retrieval scores to be directly incorporated into the final representation, ensuring the differentiability of the framework.

\subsubsection{Visual Understanding}

To enhance retrieval-based computer vision tasks, we integrate {\ours} with CLIP's architecture. In contrast to CLIP, which requires training the entire model using SGD, {\ours} focuses on training the prompt alone by constructing a query-key knowledge-store from few-shot supervisions to obtain prompt weights.
To achieve this, {\ours} employs CLIP's visual encoder to extract visual attributes from a limited training dataset of images and transforms the associated labels into one-hot encoded vectors.
 Subsequently, the knowledge-store is established, encompassing key-value combinations of visual attributes and one-hot encoded labels sourced from the limited training dataset.
During inference,  the probability distribution associated with a test image's fetched feature melds with its inherent feature encoded by CLIP. Through this fusion, {\ours} capitalizes on insights from the pre-established CLIP model as well as the limited-instance training data. Interpreting the prompt with these parameters can be perceived as tapping into few-shot wisdom from the accumulated knowledge-store.


\section{Experiment Implementation}

\begin{table*}[ht]
\centering
\small
\caption{
We report performance in both zero-shot and few-shot settings for nine NLU datasets. Our data reflects the standard and mean deviation from experiments across these datasets, averaging over five distinct few-shot splits. In our terminology, ``D-demo'' alludes to discrete demonstrations, while ``KnPr'' stands for KnowPrompt.
LOTClass~\cite{meng2020text} is acknowledged as the cutting-edge approach for unsupervised text categorization using self-training. The marker {\dag} signifies models that tap into \textbf{supplementary information}, and {$^\clubsuit$} denotes models that \textbf{distill} the PFM using the full unlabeled dataset. Contrarily, our method, and the comparative baselines exclusively utilize the standard PFM for evaluation without additional training. Average results highlighted by {$^*$} convey that we've adopted outcomes from the ``non-demo'' variant of the corresponding model to provide default results.
}
\scalebox{0.8}{
\begin{tabular}{l|l|lll|lll|l|lll|l}
\toprule
{\multirow{3}{*}{\textbf{St.}}} 
& {\multirow{3}{*}{\textbf{Model}}} 
& \multicolumn{3}{c|}{\textbf{Individual Sentence}}
& \multicolumn{3}{c|}{\textbf{A pair of Sentence }}
& {\multirow{3}{*}{\textbf{Model}}} 
& \multicolumn{3}{c|}{\textbf{Information Extraction }}
& {\multirow{3}{*}{\textbf{Avg.}}} \\
\cmidrule{3-8}
\cmidrule{10-12}

& &SST-2  & MR   &CR    &MNLI   &QNLI   &QQP  &  &FewN  &SemEval  &TACRED  \\
& & (acc)  & (acc)  & (acc)   & (acc)   & (acc)  & (F1) &   & (acc)   & (acc) & (F1)  &   \\


\midrule
    \multirow{5}{*}{16} 
   & \textsc{FT}~\  
     & 81.4 \tiny{(3.8)}
     & 76.9 \tiny{(5.9)}  
     & 75.8 \tiny{(3.2)} 
     & 45.8 \tiny{(6.4)}
   & 60.2 \tiny{(6.5)} 
   & 60.7 \tiny{(4.3)}
   & \textsc{FT}
   & 52.7 \tiny{(2.2)} 
   & 66.1 \tiny{(1.2)}
   & 25.8 \tiny{(2.8)}
   & 60.6  \\

   & \textsc{LM-BFF} (man)~\  
   & 91.6 \tiny{(1.2 )} 
   & 87.0 \tiny{(2.0)}  
   & 90.3 \tiny{(1.6)} 
   & 64.3 \tiny{(2.5)}
   & 64.6 \tiny{(5.4 )} 
   & 65.4 \tiny{(5.3)} 
   & {KnPr}~\  
   & 65.3 \tiny{(1.1)}  
   & 80.9 \tiny{(2.5)}
   & 33.2 \tiny{(2.0)}
   &  71.4 \\
   
   & \textsc{LM-BFF} (D-demo)
  & 91.8 \tiny{(1.2 )} 
  &  86.6 \tiny{(1.8)}  
  & 90.2 \tiny{(1.4)} 
  & 64.8 \tiny{(2.3)}
  & 69.2 \tiny{(5.4)}
   & 68.2 \tiny{(3.2)}
   & {KnPr} (D-demo)~
   & ~\ ~\  ---
   & ~\ ~\ ---
   & ~\ ~\ ---
   & 72.2{$^*$}  \\
  
    
 & \textsc{KPT} \dag
  & 90.3 \tiny{(1.6)}
  & 86.8 \tiny{(1.8)}  
  & 88.8 \tiny{(3.7)} 
  & 61.4 \tiny{(2.1)} 
   & 61.5 \tiny{(2.8)} 
   & 71.6 \tiny{(2.7)} 
   & \textsc{KPT} \dag 
   &  65.9 \tiny{(1.5)}  
   &  78.8 \tiny{(2.1)}  
   &  32.8 \tiny{(1.7)}
   & 70.9 \\

\cmidrule{2-13}

 & \textbf{Ours}  
 & \textbf{93.9} \tiny{(0.4)} 
 & \textbf{88.0} \tiny{(0.8)} 
 & \textbf{91.9} \tiny{(0.7)}
 & \textbf{71.1} \tiny{(1.8)}
 & \textbf{71.6} \tiny{(1.8)}
 & \textbf{74.0} \tiny{(2.0)}
 & \textbf{Ours}
 & \textbf{67.3} \tiny{(0.9)}  
 & \textbf{81.5} \tiny{(1.3)}  
 & \textbf{40.7} \tiny{(0.7)}  
 & \textbf{75.6} \\
 
 \midrule
    \multirow{5}{*}{4} 
   & \textsc{FT}~\  
   & 60.2 \tiny{(2.8)} 
    & 57.6 \tiny{(1.4)}
   & 66.4 \tiny{(5.5)}
   & 35.0 \tiny{(0.3)}
   & 54.2 \tiny{(3.9)} 
   & 52.8 \tiny{(4.7)}  
   & \textsc{FT} 
   & 32.7 \tiny{(2.9)} 
   & 38.8 \tiny{(2.0)}
   & 14.7 \tiny{(2.8)}
   & 45.8 \\
   
   & \textsc{LM-BFF} (man)~\  
   & 90.7 \tiny{(0.8)} 
   & 85.2 \tiny{(2.8)}
   & 89.9 \tiny{(1.8)} 
   & 51.0 \tiny{(2.5)}
   & 61.1 \tiny{(6.1)} 
   & 48.0 \tiny{(4.9)} 
   & {KnPr}
   &  52.5 \tiny{(1.5)}  
   &  58.4 \tiny{(3.7)}
   &  28.8 \tiny{(2.5)}
   &  62.8\\
   
   & \textsc{LM-BFF} (D-demo)~\  
   & 90.2 \tiny{(1.5)}
   & 85.5 \tiny{(2.1)}
   & 89.7 \tiny{(0.6)} 
   & 56.1 \tiny{(1.0)}
   & 61.7 \tiny{(7.6)} 
   & 63.2 \tiny{(5.6)} 
   & {KnPr} (D-demo)
   &  ~\  ---
   &  ~\  ---
   &   ~\  ---  
   &  {65.1}{$^*$}  \\

   & \textsc{KPT} \dag
   &  88.2 \tiny{(5.7)} 
   &  83.4 \tiny{(1.5)}  
   &  87.2 \tiny{(2.5)}
   &  53.7 \tiny{(2.7)}
   &  59.2 \tiny{(2.8)} 
   & 54.9  \tiny{(7.9)}
   & \textsc{KPT} \dag
   & 58.8 \tiny{(2.2)}   
   & 57.2 \tiny{(3.2)}
   & 27.5 \tiny{(2.2)}
   & 63.3 \\

\cmidrule{2-13}
     & \textbf{Ours}  
      &  \textbf{91.5} \tiny{(1.8)}
      &  \textbf{87.4} \tiny{(0.5)}
      &  \textbf{91.4} \tiny{(0.6)} 
      &  \textbf{57.6} \tiny{(5.5)}
      &  \textbf{62.2} \tiny{(6.0)}
      &  \textbf{66.1} \tiny{(4.1)}
      & \textbf{Ours}
      &  \textbf{60.9} \tiny{(1.9)} 
      & \textbf{59.2} \tiny{(3.0)} 
      & \textbf{32.1} \tiny{(2.0)}
      & \textbf{67.6} \\
 
\midrule
    \multirow{7}{*}{0} 

   & {LOTClass}$^\clubsuit$ ~\  
   &  71.8
   &  81.7
   &  50.1
   &  50.4
   &  36.5
   & 55.9
   & {LOTClass}$^\clubsuit$ ~\ 
   & 11.5
   & 9.8
   & 2.5
   &  41.1 \\
   
   & \textsc{FT}~\  
   &  49.1
   &  50.0
   &  49.8
   &  34.4
   &  49.5
   &  31.6
   & {FT}
   & 10.0
   & 6.2
   & 0.5
   & 31.2 \\
   
   & \textsc{LM-BFF} (man)~\  
   & 83.5  
   & 80.3
   & 78.4
   & 49.7
   & 50.5
   & 49.7
   & {KnPr} 
   & 15.9 
   & 10.3
   & 2.3
   & 46.7 \\
   
     & \textsc{LM-BFF} (D-demo)~\  
     & 82.9 
     & 80.7
     & \textbf{81.4}
     & 52.2
     & 53.5
     & 44.0
     & {KnPr} (D-demo)
     &  ~\  ---
     &  ~\  ---
     &  ~\  ---  
     & 47.0{$^*$} \\
    
   & \textsc{KPT} \dag
   &  78.4
   &  81.9
   & 71.4
   &  37.1
   &  55.3
   &  47.5
   & \textsc{KPT} \dag
   &  24.6
   & 11.6
   &  0.8
   & 45.7 \\

\cmidrule{2-13}
      & \textbf{Ours} 
      &  \textbf{86.8}
      &  \textbf{83.5}
      &  {79.7}
      &  \textbf{53.7}
      &  \textbf{56.2}
      &  \textbf{56.7}
     & \textbf{Ours} 
      &  \textbf{41.3}
      & \textbf{12.2}
      & \textbf{2.8}
      &  \textbf{52.5} \\
\bottomrule

\end{tabular}
}
\label{tab:experiment-few-shot}
\end{table*}


\begin{figure*}[htp]
    \centering
\includegraphics[width=0.88\textwidth]{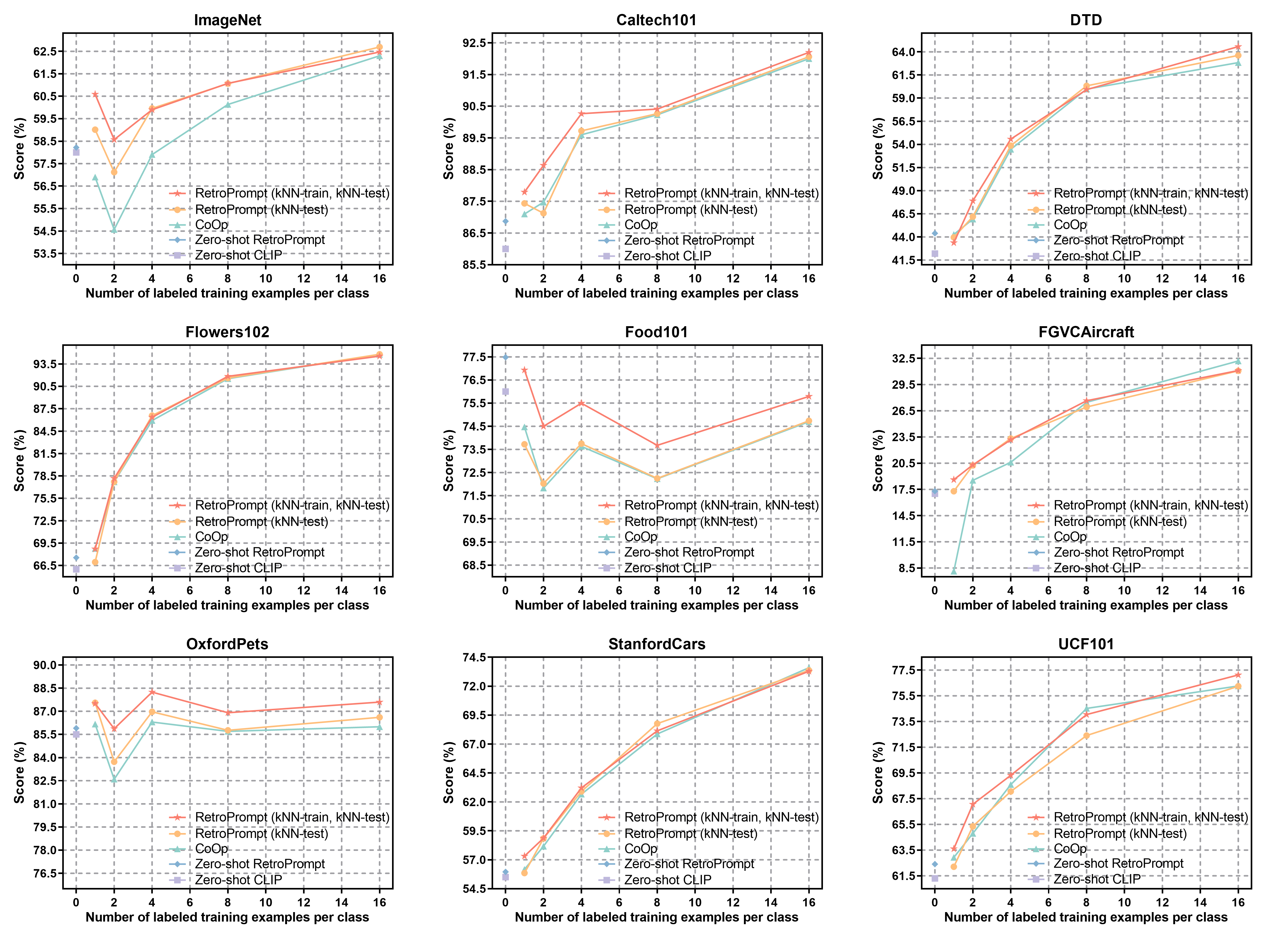}
\caption{
We present the results on 9 image classification datasets in both the zero-shot and few-shot settings.
For the few-shot setting, we employ \knn{}-train, which involves retrieving \knn{} to guide the training process. On the other hand, for the zero-shot setting, we utilize \knn{}-test, where \knn{} is retrieved to interpolate predictions.}
\label{fig:cv_results}
\end{figure*}

\subsection{Datasets}

\textbf{Language Understanding.}
We assess our approach on various natural language tasks, including single sentence classification (e.g. SST-2~\cite{sst2}, MR~\cite{mr}, CR~\cite{cr}) and sentence pairs (MNLI~\cite{mnli}, QNLI~\cite{qnli}, QQP\footnote{\url{https://www.quora.com/q/quoradata}}). Additionally, to evaluate multi-class capability, we experiment on information extraction datasets like  SemEval \cite{hendrickx2010semeval}, TACRED \cite{DBLP:conf/emnlp/ZhangZCAM17},
 and FewNERD~\cite{fewnerd}. 

\hspace*{\fill} \\
\noindent
\textbf{Visual Understanding.}
We perform experiments on {\ours} using 9 publicly available image classification datasets, namely ImageNet~\cite{DBLP:conf/cvpr/DengDSLL009}, Caltech101~\cite{DBLP:conf/cvpr/LiFP04}, DTD~\cite{DBLP:conf/cvpr/CimpoiMKMV14}, FGVCAircraft~\cite{DBLP:journals/corr/MajiRKBV13}, Flowers102~\cite{DBLP:conf/icvgip/NilsbackZ08}, Food101~\cite{DBLP:conf/eccv/BossardGG14}, OxfordPets~\cite{DBLP:conf/cvpr/ParkhiVZJ12}, Stanford Cars~\cite{DBLP:conf/iccvw/Krause0DF13}, and UCF101~\cite{DBLP:journals/corr/abs-1212-0402}.
These datasets encompass various vision tasks, including generic object classification, action recognition, fine-grained category classification, as well as  texture recognition.

\subsection{Baselines}
\textbf{Language Understanding.}
We compare our approach to prior methods on single and multi-sentence tasks. For classification, baselines include LM-BFF~\cite{gao2020making} and KnowPrompt~\cite{chen21knowprompt}, a leading prompt tuning model. For information extraction, KnowPrompt serves as the baseline given its demonstrated efficacy.
As discrete demonstration cannot accommodate multi-class inputs, we exclude results for KnowPrompt using demonstrations. Additionally, we compare our {\ours} to KPT~\cite{KPT}, an approach enhancing prompting with external knowledge bases unlike our focus on leveraging internal training data as the knowledge source.

\hspace*{\fill} \\
\noindent
\textbf{Visual Understanding.}
In the domain of visual understanding, 
We juxtapose {\ours} against two foundational techniques leveraging expansive pre-trained vision-language architectures. The inaugural benchmark is the zero-shot CLIP~\cite{clip}. This strategy classifies by determining affinities between the visual characteristics of test illustrations and the textual attributes of meticulously crafted cues. Notably, it operates devoid of supplementary training specimens. For standard entities and backdrops, the cue design is ``an image showcasing a [CLASS].'' In scenarios of nuanced categories, pertinent context is infused: e.g., ``a specific pet breed'' for OxfordPets and ``a distinct culinary item" for Food101. The secondary benchmark is CoOp~\cite{DBLP:journals/ijcv/ZhouYLL22}, which interprets the contextual terminology of a cue via adaptable vectors, all while maintaining the integrity of the pre-trained coefficients.

\subsection{Evaluation Protocols and Details}
\label{subsec:details}

\subsubsection{Language Understanding}

Our study is developed using PyTorch for the machine learning framework and evaluated on one NVIDIA Tesla V100 GPU with 32GB memory.  
We employ $\text{RoBERTa}_\text{large}$ as the PFM and select AdamW for optimization across all tests. To ensure consistent evaluation, we use the same templates for both the baselines and {\ours} across each dataset, mitigating the influence of diverse templates. We perform experiments in two different settings: zero-shot and few-shot.

\hspace*{\fill} \\
\noindent
\textbf{Few-shot Setting}
We adopt the methodology from LM-BFF~\cite{gao2020making} and perform evaluations under 4-shot and 16-shot configurations. Performance is gauged by averaging results across diverse sampled $\dtrain$ for each task, using a consistent set of seeds, denoted as $\seedset$. 
It's crucial to highlight that the knowledge-store is derived from the \textbf{few-shot training set} in these scenarios.

\hspace*{\fill} \\
\noindent
\textbf{Zero-shot Setting\footnote{Note that it does not strictly fall under the category of zero-shot sense.}}
In the zero-shot configuration, we employ the standard $\text{RoBERTa}_\text{large}$ for direct testing on the dataset, barring the LOTClass~\cite{meng2020text} approach.
To exploit retrieval, our approach follows LOTClass~\cite{meng2020text} utilizing unlabeled training data. 
Specifically, $\text{RoBERTa}_\text{large}$ assigns pseudo-labels to unlabeled train data, constructing an open-source knowledge-store of pseudo-labeled instances. Finally, predictions are issued on the test set utilizing this knowledge \textbf{without fine-tuning parameters}, aligned with true zero-shot evaluation.

\subsubsection{Visual Understanding}
Experiments were carried out utilizing a singular NVIDIA GeForce RTX 3090 24G GPU. We use the default version of CoOp, where the class token is positioned at the end and a unified context is learned as the foundation of our framework.
Unless otherwise specified, we leverage ResNet-50~\cite{resnet} as the foundational architecture for our image encoder and allocate a capacity of 16 for context tokens. All architectural designs stem from the publicly accessible source code of CLIP\footnote{\url{https://github.com/openai/CLIP}.}.

\hspace*{\fill} \\
\noindent
\textbf{Few-shot Setting.}
We adopt  the few-shot configuration of CLIP~\cite{clip}, training with 1, 2, 4, 8, and 16 shot instances and subsequently testing the models on comprehensive test datasets. The outcomes presented are averaged across three iterations for a standardized comparison. It's worth mentioning that in this scenario, the knowledge-store is assembled from the \textbf{few-shot training set}.

\hspace*{\fill} \\
\noindent
\textbf{Zero-shot Setting.}
Similar to the setting in NLU experiments, {\ours} leverages the unlabeled training set for retrieval to benefit from the retrieval mechanism.
Specifically, we employ zero-shot CLIP on the pseudo-labeled training set to construct our datastore. 
Our method, {\ours}, forecasts the outcomes on the test set using the built datastore, \textbf{bypassing any parameter adjustments}. The results we showcase are averaged across three iterations for a consistent comparison.

\section{Experimental Results}

In the ensuing section, we delineate the detailed outcomes from {\ours} and juxtapose them against foundational models under the purview of both zero-shot and few-shot scenarios. We further provide insights into the performance of {\ours}.

\subsection{Few-shot Results}

\textbf{Language Understanding.}
Table~\ref{tab:experiment-few-shot} highlights the consistent edge that {\ours} holds over benchmark models such as KnowPrompt and LM-BFF  in the 4-shot and 16-shot studies. When tasked with multi-class information extraction, conventional discrete demonstrations often falter due to restrictions on input sequence length. On the other hand, our neural demonstration adeptly navigates such challenges, leading to enhancements in multi-class dataset performance. Moreover, {\ours} surpasses KPT, delivering superior results without the crutch of external knowledge — our approach relies strictly on the in-house few-shot datasets. Another noteworthy observation is the reduced standard deviation manifested by {\ours} in comparison to the reference models, suggesting that our retrieval methodology effectively mitigates variability in parametric estimations.

\hspace*{\fill} \\
\noindent
\textbf{Visual Understanding.}
We introduce two versions of our method, {\ours}, both demonstrating excellent performance compared to other approaches (Figure~\ref{fig:cv_results}). The default version uses \knn{} for training guidance and prediction interpolation, while the second employs \knn{}-based probability. With limited training instances, {\ours} consistently outperforms CoOp, eliminating the need for labor-intensive fine-tuning. For example, on the FGVCAircraft dataset, {\ours} surpasses CoOp by 10.48\% and 1.78\% in 1-shot and 2-shot scenarios, respectively. Similarly, on Food101, {\ours} achieves 2.46\% and 2.67\% improvements in the same settings. Notably, using just a single shot, {\ours} significantly outperforms zero-shot CLIP on average. When trained with 16 shots, {\ours} surpasses zero-shot CLIP by approximately 13\% on average.

\subsection{Zero-shot Results}
\textbf{Language Understanding.}
As shown in Table~\ref{tab:experiment-few-shot}, {\ours} showcases enhancement. Notably, {\ours} surpasses KPT under these conditions, suggesting that leveraging inherent data to separate knowledge from rote learning offers more promise than depending on external insights. Additionally, our methodology outshines LOTClass, even when deploying the standard $\text{RoBERTa}_\text{large}$ devoid of extra training.

\hspace*{\fill} \\
\noindent
\textbf{Visual Understanding.}
Figure~\ref{fig:cv_results} affirms our observations. Even without training, {\ours} continually outperforms zero-shot CLIP under the zero-shot scenario. This unwavering advantage over the nine datasets underscores {\ours}'s efficacy and universal zero-shot applicability.

\begin{figure} 
\centering 
\includegraphics[width=0.28\textwidth]{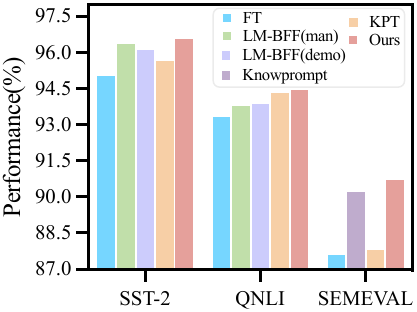} 
\caption{Fully-supervised performances.}
\label{fig:fully_supervised}
\vspace{-0.5cm}
\end{figure}

\begin{table}[ht]
\centering
\caption{Model generalization results in Language Understanding tasks.}
\scalebox{0.85}{
\begin{tabular}{l|c|cc}
\toprule
 {\multirow{1}{*}{\textbf{Model}}} 

& \multicolumn{1}{c|}{\textbf{Source}}
& \multicolumn{2}{c}{\textbf{Target Domain}}
\\
\cmidrule{1-4}
&16-shot MR &SST-2 & CR   \\

\midrule

    \textsc{FT}~\  
   & 76.9
   & 71.4
   & 64.7

  \\

    \textsc{LM-BFF} (D-demo)~\  
     & 86.6 & 89.3  & 87.5
\\

    \textsc{LM-BFF} (man)~\  
   & 87.0
   & 88.9
   & 86.9
 \\
    
  \textsc{KPT} 
   & 86.8 & 86.8  & 86.7
\\

\cmidrule{1-4}
     \textbf{\ours}  
      &  \textbf{88.0}  &   \textbf{91.4}  &   \textbf{88.8}
\\
\cmidrule{1-4}
 &16-shot QQP &MRPC & RTE   \\

\midrule
    \textsc{FT}~\  
   & 60.7
   & 43.7
   & 48.0

  \\

    \textsc{LM-BFF} (D-demo)~\  
     & 68.2
     & 38.8
     & 66.2
\\
    \textsc{LM-BFF} (man)~\  
   & 65.4
   & 20.9
   & 65.5
 \\
    
  \textsc{KPT} 
   & 71.6
   & 42.3
   & 65.8
\\

\cmidrule{1-4}
     \textbf{\ours}  
      &  \textbf{74.0}  
      & \textbf{49.4}  &   \textbf{67.3}
 \\
\bottomrule
\end{tabular}
}
\label{tab:experiment-cross-domain}
\end{table}

\subsection{Fully-supervised Results} Figure~\ref{fig:fully_supervised} illustrates that in fully-supervised scenarios with long-tail distributions, {\ours} consistently excels over the benchmark models. This suggests that our retrieval approach bolsters the PFM's proficiency in assimilating complex instances within such datasets.

\begin{table}[htp!]
\centering
\caption{Results from the model's domain adaptation within Visual Comprehension tasks.}
\scalebox{0.85}{
\begin{tabular}{l|c|cccc}
\toprule
 {\multirow{1}{*}{\textbf{Model}}} 

& \multicolumn{1}{c|}{\textbf{Source}}
& \multicolumn{4}{c}{\textbf{Target Domain}}
\\
\cmidrule{1-6}
& ImageNet & -V2 & -Sketch & -A & -R \\

\midrule
    
  \textsc{Zero-Shot CLIP} 
 & 58.18 & 51.34 & \textbf{33.32} & 21.65 & \textbf{56.00}
\\

\midrule
    
  \textsc{CoOp} 
  & 62.30 & 55.11 & 32.74 & 22.12 & 54.96
\\

\cmidrule{1-6}
   \textbf{\ours}  
      & \textbf{62.65} 
      & \textbf{55.49}  
      & 32.89
      & \textbf{23.29} 
      & 55.31 \\
\bottomrule
\end{tabular}
}
\label{tab:cv-experiment-cross-domain}
\end{table}

\subsection{Generalization of the Model to Novel Domains}

\textbf{Language Understanding.}
The presence of limited data may lead to overfitting issues for the memory parameters of PFMs, despite prompt learning techniques.
To validate the generalization capability of the {\ours}, we perform cross-domain experiments by training our model on the source datasets and subsequently evaluating its performance on diverse target datasets.
Table~\ref{tab:experiment-cross-domain} demonstrates that our method consistently outperforms the baselines, showcasing the excellent model generalization ability of {\ours} to new domains.

\hspace*{\fill} \\
\noindent
\textbf{Visual Understanding.}
In line with CoOp~\cite{DBLP:journals/ijcv/ZhouYLL22}, we perform domain generalization tests utilizing ImageNet~\cite{DBLP:conf/cvpr/DengDSLL009} as the primary dataset and employ four distinct ImageNet variations, each with unique domain deviations, as the evaluation datasets.
As presented in Table~\ref{tab:cv-experiment-cross-domain}, {\ours} enhances the robustness of CLIP to distribution shifts, and the learned prompts demonstrate generalizability.
Furthermore, the retrieval mechanism effectively improves the transfer learning capability of pre-trained foundation models on target datasets with minimal effort in constructing the datastore.

\subsection{Analysis of Memorization}
\label{subsec:memorization}

Conducting an in-depth analysis of the memorization mechanism is crucial and intriguing as it facilitates a deeper grasp of the impact of retrieval in NLP memorization.

\hspace*{\fill} \\
\noindent
\textbf{Formulation of Memorization Metrics.}
Motivated by the insights from \cite{feldman2020does} in the domain of CV, we introduce \textit{memorization measures} to evaluate the influence when one training sample $\bm{a}$ is omitted from the training set. In alignment with the methodologies presented in \cite{koh2017understanding,memory}, we conceptualize and compute the memorization score for a training sample $\bm{a}$ in a subsequent manner:

\begin{equation}
\small
\label{equ:remove}
\begin{aligned}
    {\mathcal{S}}_{\text{delate}}(\bm{a}) 
    &\overset{\text{def}}{=} -\frac{d P(y|\bm{x}; \Hat{\theta}_{\xi, -\bm{a}})}{d \xi} \bigg|_{\xi=0} \\
    &= -\nabla_{\theta}P(y|\bm{x}; \hat{\theta})^{\top}\frac{d \hat{\theta}_{\xi, -\bm{a}}}{d \xi} \bigg|_{\xi=0} \\
    &= -\nabla_{\theta}P(y|\bm{x}; \hat{\theta})^{\top}H^{-1}_{\hat{\theta}}\nabla_{\theta}{\mathcal{L}(\bm{a}, \hat{\theta})},
\end{aligned}
\end{equation}
 $\hat{\theta}_{\xi, -\bm{a}}$ signifies the parameters adjusted by down-weighting the instance $\bm{a}$ by a factor of $\xi$. $\hat{\theta}$ corresponds to the model parameters trained considering all instances. 
This term $H_{\hat{\theta}}$  is operationalized as the mean of the second-order partial derivatives of the loss objective regarding the parameters, given by $\frac{1}{n}\sum^{n}_{i=1}{\nabla^{2}_{\theta}{\mathcal{L}(a_i, \hat{\theta})}}$. The measure $\mathcal{S}_{\text{delete}}(\bm{a})$ effectively gauges the extent of variation in $P(y|\bm{x}; \theta)$ upon down-weighting the instance $\bm{a}$ by $\xi$.

\begin{table}[ht]
\centering
\caption{The upper section displays the mean percentage of \emph{positive phrases} across various memory groups of negative/positive instances.
The lower part denotes the mean values of memorization score on the SST-2 dataset.
}
\label{table.atypical}
\begin{small}
\scalebox{0.8}{

\begin{tabular}{l|ccc|ccc}
\toprule
\multirow{2}{*}{\textbf{Mem Group}} & \multicolumn{3}{c}{\textbf{Negative}} & \multicolumn{3}{c}{\textbf{Postive}} \\

\cmidrule{2-7}
&FT  & LM-BFF & OURS 
&FT  & LM-BFF & OURS \\

\midrule
Top-10\%  &34.29  & 32.78  & 30.23
          & 68.75  & 69.71 &75.67 \\
ALL                        & \multicolumn{3}{c}{23.40}      & \multicolumn{3}{c}{86.39}        \\

Bottom-10\% & 17.63 & 16.25 & 14.42
& 95.92 & 95.08 & 94.53 
\\

\bottomrule
\midrule

 & \multicolumn{2}{c}{FT} & \multicolumn{2}{c}{LM-BFF} & \multicolumn{2}{c}{OURS} \\
\midrule

\textsc{Mem Score}            & \multicolumn{2}{c}{4.597}   & \multicolumn{2}{c}{0.121}       & \multicolumn{2}{c}{0.032}  \\
\bottomrule
\end{tabular}
}
\end{small}
\vspace{-0.4cm}
\end{table}

\hspace*{\fill} \\
\noindent
\textbf{Top-memorized Samples: Typical or Atypical?}
We analyze memorization on the SST-2 dataset, which supplies phrase-level sentiment labels. Atypicality is assessed through the proportion of positive phrases in an instance.
Computations on SST-2 reveal typical positive instances usually contain a high percentage of positive phrases, whereas typical negative instances contain few. Leveraging this, The memorization score specified in Equation~\ref{equ:remove} is utilized in our methodology 
 to choose the bottom and top 10\% memorized training instances and compute the mean proportion of the positive phrases within each group.

\begin{table}[ht]
\centering
\caption{
We conduct detailed ablation experiments in few-shot settings for Language Understanding tasks.
The notation ``N-demo'' represents the neural demonstration, whereas ``refresh'' pertains to the knowledge-store's asynchronous update.
}
\scalebox{0.9}{
\begin{tabular}{l|ccccc}
\toprule
\multirow{2}{*}{\textbf{Model}}                      
& \multicolumn{5}{c}{\textbf{16-shot}} 
\\ 
\cmidrule{2-6} 
 & TACRED  & CR & SST-2   & MNLI & QQP   
  \\ \midrule
\textbf{\ours}
& \textbf{40.7}
& \textbf{91.9}
& \textbf{93.9}  
 & \textbf{71.1}
& \textbf{74.0}      
    \\

\midrule
w/o \text{\knn{}}-test
& 38.2
& 91.2
& 93.2 
& 70.4
&  73.0      \\
w/o \text{\knn{}}-train
& 36.5
& 90.2
& 92.0
& 68.8
&  71.3       \\
w/o N-demo
& 37.9
& 91.0 
& 92.4
& 70.1
&  72.7       \\ 
w/o refresh
& 39.9
& 91.5
& 93.5
& 70.7
&  73.6      \\ 
\bottomrule
\end{tabular}}
\label{tab:ablation}
\end{table}

\begin{table}[ht]
\centering
\caption{Detailed ablation experiments in few-shot settings in Visual Understanding tasks. 
}
\scalebox{0.8}{
\begin{tabular}{l|ccccc}
\toprule
\multirow{2}{*}{\textbf{Model}}                      
& \multicolumn{5}{c}{\textbf{16-shot}} 
\\ 
\cmidrule{2-6} 
  & Caltech101 & DTD & FGVCAircraft & Food101  & OxfordPets
  \\ \midrule
\textbf{\ours}
& \textbf{92.20} 
& \textbf{64.56} 
& \textbf{31.10} 
& \textbf{75.79} 
& \textbf{87.59} \\
\midrule
w/o \text{\knn{}}-test
& 91.74
& 63.73
& 30.63
& 75.77
& 87.50 \\
w/o \text{\knn{}}-train
& 92.06
& 63.61
& 31.08
& 74.74
&  86.61    \\
w/o refresh
& 91.66
& 64.3
& 31.04
& 75.67
& 87.01    \\ 
\bottomrule
\end{tabular}}
\label{tab:cv-ablation}
\end{table}

As depicted in Table~\ref{table.atypical}, we draw the following conclusions from our findings:
(1) The PFM exhibits a tendency to allocate deeper memory to atypical samples. Both the LM-BFF approach and our technique have shown that the top decile of memorized negative samples contain a greater proportion of positive phrases relative to the mean percentage found across all negative instances.
(2) Compared to fine-tuning, LM-BFF displays diminished memorization when dealing with complex instances.
The observed phenomenon can be ascribed to the \textbf{prompting that allows PFMs to tap into pre-trained knowledge without amplifying recall for subsequent data}. (3) The average memorization scores for {\ours} are notably lower than those observed for fine-tuning and LM-BFF. This suggests our approach's diminished reliance on recall. The underlying reason for this is the \textbf{strategy of disentangling knowledge from sheer recall via retrieval, thereby curbing the inherent propensities of PFMs for verbatim memorization}.

\subsection{Case Analysis}
\begin{figure} 
\centering 
\includegraphics[width=0.5\textwidth]{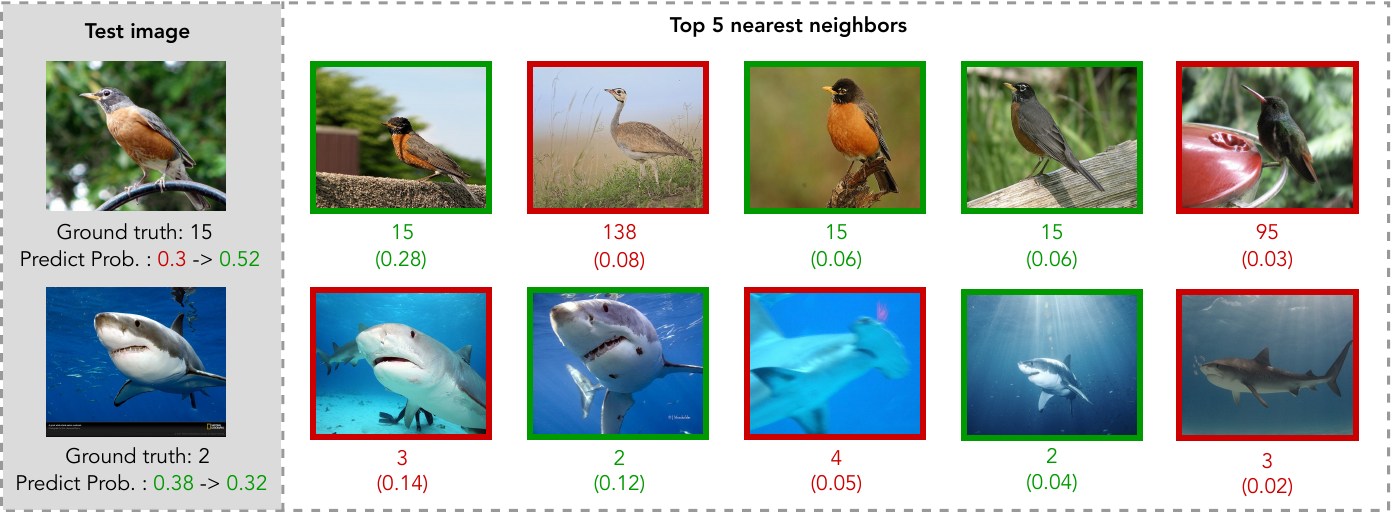} 
\caption{Case examples of Top-5 neighbors from trainset of ImageNet. The numbers before and after the arrow in ``Predict Prob'' represent the values before and after retrieval.}
\label{fig:case}
\end{figure}

For a deeper understanding of \knn{}'s function, we undertake a meticulous review of ImageNet instances, illustrated in Figure~\ref{fig:case}. For each case, we visualize the top 5 nearest neighbors and their respective \knn{} probabilities. Our findings indicate that \knn{} significantly improves the accuracy of incorrect predictions. Moreover, even in cases where \knn{} predictions fail, the impact on correct predictions is minimal, as the probability of the ground truth in the \knn{} distribution remains high.

\subsection{Ablation Studies}
\label{ablation}
\textbf{Component Ablation}
As depicted in Table~\ref{tab:ablation} and Table~\ref{tab:cv-ablation}, the outcomes from the four distinct component ablation experiments register a pronounced decline, underscoring the vital role of our retrieval component. Importantly, the enhancements from \knn{}-train and neural demonstration appear more pronounced in few-shot scenarios compared to those observed in \knn{}-test.
It's noteworthy that \knn{}-test, akin to \knn{}-LM~\cite{DBLP:conf/iclr/KhandelwalLJZL20,he2021efficient}, exerts limited impact when integrated solely within the testing phase of prompt learning.

\begin{table}[ht]
    \caption{
     Results on the 16-shot CR and TACRED datasets, considering various key attributes and methodologies for computing the \knn{}.}
    \label{result-ktype}
    \centering
    \scalebox{0.9}{
    \begin{tabular}{llcc}
    \toprule
    Key Feat. &  \knn{} Sel. & CR & TAC.  \\
    \midrule
    Prompt & Feat-similar & 91.9 & 40.7\\
    \texttt{[CLS]}  & Feat-similar & 89.0 & 37.2    \\      
    Prompt  & BM25 & 89.5 & 38.8 \\
    \texttt{[CLS]} & BM25 & 88.7 & 36.1 \\ 
    \bottomrule
    \end{tabular}
    }
\end{table}

\hspace*{\fill} \\
\noindent
\textbf{Key Feature and \knn{} Selection}
We investigate the impact of key feature type and \knn{} selection method in our knowledge store. For key features, we compare: (1) prompt-based (default), and (2) [CLS]-based on the language model. For \knn{}  distribution calculation, we examine (1) feature similarity score (referred to as feat-similar, default), and (2) the BM25-derived metric computes the affinity score between a given query and individual key instances, leveraging the BM25 algorithm. While Table~\ref{result-ktype}  indicates prompt-based key features with feature similarity scores for \knn{}  yield the best performance. This implies prompts facilitate more effective context representations for similarity assessments, outperforming BM25-based scores.

\section{Conclusion}
In this study, we introduced {\ours}, a novel approach that enhances the generalization ability of prompt learning for pre-trained foundation models by decoupling knowledge from memorization through retrieval augmentation. We successfully applied the \knn{} guider to both textual and visual comprehension tasks, leading to superior performance in zero-shot, few-shot, and fully-supervised settings compared to other prompt learning and knowledge-augmented prompt methods. Our analysis confirms the effectiveness of decoupling knowledge from memorization in achieving better results.
However, this approach introduces computational overhead from its retrieval operations and faces scalability challenges when applied to massive foundation models such as GPT-4 and LLaMA.
Future work could extend RETROPROMPT to generative applications, such as image captioning, and explore its effectiveness in multilingual settings to broaden its utility.

\section*{Acknowledgments}
 
This work was supported by the National Natural Science Foundation of China (Nos. 62506166, U2441285, 62206246, 62222605),  the Natural Science Foundation
of Jiangsu Province (No. SBK20250401456), the China Postdoctoral Science Foundation (No. 2025M774283), the Yongjiang Talent Introduction Programme (No. 2021A-156-G) and
the Ningbo Natural Science Foundation (No. 2024J020).
This research is also sponsored by the DiDi GAIA Collaborative Research Funds (No. CCF-DiDi GAIA202507), and Information Technology Center and State Key Lab of CAD\&CG, Zhejiang University.




\bibliographystyle{IEEEtranN}
\bibliography{reference}

@article{radfordimproving,
  title={Improving Language Understanding by Generative Pre-Training},
  author={Radford, Alec and Narasimhan, Karthik and Salimans, Tim and Sutskever, Ilya}
}

@inproceedings{resnet,
  author    = {Kaiming He and
               Xiangyu Zhang and
               Shaoqing Ren and
               Jian Sun},
  title     = {Deep Residual Learning for Image Recognition},
  booktitle = {
               {CVPR} 2016},
  pages     = {770--778},
  publisher = {{IEEE} Computer Society},
  year      = {2016},
}

@inproceedings{vit,
  author    = {Alexey Dosovitskiy and
               Lucas Beyer and
               et al.},
  title     = {An Image is Worth 16x16 Words: Transformers for Image Recognition
               at Scale},
  booktitle = { {ICLR} 2021},
  publisher = {OpenReview.net},
  year      = {2021},
}

@inproceedings{clip,
  author    = {Alec Radford and
               Jong Wook Kim and
               et al.},
  title     = {Learning Transferable Visual Models From Natural Language Supervision},
  booktitle = {
               {ICML} 2021},
  series    = {Proceedings of Machine Learning Research},
  volume    = {139},
  pages     = {8748--8763},
  publisher = {{PMLR}},
  year      = {2021}
}

@inproceedings{
chen2022decoupling,
title={Decoupling Knowledge from Memorization: Retrieval-augmented Prompt Learning},
author={Xiang Chen and Lei Li and Ningyu Zhang and Xiaozhuan Liang and Shumin Deng and Chuanqi Tan and Fei Huang and Luo Si and Huajun Chen},
booktitle={Advances in Neural Information Processing Systems},
year={2022},
}

@inproceedings{DBLP:conf/emnlp/ZhangZCAM17,
  author    = {Yuhao Zhang and
               Victor Zhong and
               Danqi Chen and
               Gabor Angeli and
               Christopher D. Manning},
  title     = {Position-aware Attention and Supervised Data Improve Slot Filling},
  booktitle = {Proc. of {EMNLP} 2017},
  year      = {2017},
}

@inproceedings{hendrickx2010semeval,
    title = "{S}em{E}val-2010 Task 8: Multi-Way Classification of Semantic Relations between Pairs of Nominals",
    author = "Hendrickx, Iris  and
      Kim, Su Nam  and et al.",
    booktitle = "Proceedings of SemEval",
    url = "https://www.aclweb.org/anthology/S10-1006/",
    year = "2010",
    pages = "33--38"
}

@article{KPT,
  author    = {Shengding Hu and
               Ning Ding and
               Huadong Wang, et al},
  title     = {Knowledgeable Prompt-tuning: Incorporating Knowledge into Prompt Verbalizer
               for Text Classification},
  journal   = {ACL },
  year      = {2022}
}

@inproceedings{DBLP:conf/nips/BrownMRSKDNSSAA20,
  author    = {Tom B. Brown and
               Benjamin Mann, etal},
  title     = {Language Models are Few-Shot Learners},
  booktitle = {Proceedings  of NeurIPS 2020},
  year      = {2020},

}

@inproceedings{DBLP:conf/nips/LewisPPPKGKLYR020,
  author    = {Patrick S. H. Lewis and
               Ethan Perez and
               et al.},
  title     = {Retrieval-Augmented Generation for Knowledge-Intensive {NLP} Tasks},
  booktitle = {NeurIPS 2020},
  year      = {2020}
}

@inproceedings{RETRO,
  author    = {Sebastian Borgeaud and
               Arthur Mensch, etal},
  title     = {Improving Language Models by Retrieving from Trillions of Tokens},
  booktitle = {{ICML} 2022},
  publisher = {{PMLR}},
  year      = {2022},

}

@article{DBLP:journals/corr/abs-2112-08633,
  author    = {Ohad Rubin and
               Jonathan Herzig and
               Jonathan Berant},
  title     = {Learning To Retrieve Prompts for In-Context Learning},
  journal   = {CoRR},
  volume    = {abs/2112.08633},
  year      = {2021},

}

@article{DBLP:journals/corr/abs-2101-06804,
  author    = {Jiachang Liu and
               Dinghan Shen and
               Yizhe Zhang and
               Bill Dolan and
               Lawrence Carin and
               Weizhu Chen},
  title     = {What Makes Good In-Context Examples for GPT-3?},
  journal   = {CoRR},
  volume    = {abs/2101.06804},
  year      = {2021},
}

@inproceedings{sst2,
  author    = {Richard Socher and
               Alex Perelygin and
               et al.},
  title     = {Recursive Deep Models for Semantic Compositionality Over a Sentiment
               Treebank},
  booktitle = { {EMNLP} 2013},
  publisher = {{ACL}},
  year      = {2013},
 
}

@inproceedings{cr,
  author    = {Minqing Hu and
               Bing Liu},
  title     = {Mining and summarizing customer reviews},
  booktitle = { {SIGKDD}, 2004},
  pages     = {168--177},
  publisher = {{ACM}},
  year      = {2004},
}

@inproceedings{mr,
  author    = {Bo Pang and
               Lillian Lee},
  title     = {Seeing Stars: Exploiting Class Relationships for Sentiment Categorization
               with Respect to Rating Scales},
  booktitle = {{ACL} 2005},
  pages     = {115--124},
  publisher = {The Association for Computer Linguistics},
  year      = {2005},
}

@inproceedings{mnli,
  author    = {Adina Williams and
               Nikita Nangia and
               Samuel R. Bowman},
  title     = {A Broad-Coverage Challenge Corpus for Sentence Understanding through
               Inference},
  booktitle = {
               {NAACL-HLT} 2018},
  pages     = {1112--1122},
  publisher = {Association for Computational Linguistics},
  year      = {2018},

}

@inproceedings{qnli,
  author    = {Pranav Rajpurkar and
               Jian Zhang and
               Konstantin Lopyrev and
               Percy Liang},
  title     = {SQuAD: 100, 000+ Questions for Machine Comprehension of Text},
  booktitle = {{EMNLP} 2016},
  year      = {2016}
}

@inproceedings{fewnerd,
  author    = {Ning Ding and
               Guangwei Xu and
               Yulin Chen and
               Xiaobin Wang and
               Xu Han and
               Pengjun Xie and
               Haitao Zheng and
               Zhiyuan Liu},
  title     = {Few-NERD: {A} Few-shot Named Entity Recognition Dataset},
  booktitle = {{ACL/IJCNLP} 2021},
  pages     = {3198--3213},
  publisher = {Association for Computational Linguistics},
  year      = {2021}
}

@inproceedings{DBLP:conf/iclr/KhandelwalFJZL21,
  author    = {Urvashi Khandelwal and
               Angela Fan and
               Dan Jurafsky and
               Luke Zettlemoyer and
               Mike Lewis},
  title     = {Nearest Neighbor Machine Translation},
  booktitle = { {ICLR} 2021},
  publisher = {OpenReview.net},
  year      = {2021},
}

@inproceedings{focal_loss,
  author    = {Tsung{-}Yi Lin and
               Priya Goyal and
               Ross B. Girshick and
               Kaiming He and
               Piotr Doll{\'{a}}r},
  title     = {Focal Loss for Dense Object Detection},
  booktitle   = {{IEEE} Trans. Pattern Anal. Mach. Intell.},
  volume    = {42},
  pages     = {318--327},
  year      = {2020}
}

@inproceedings{elangovan-etal-2021-memorization,
    title = "Memorization vs. Generalization : Quantifying Data Leakage in {NLP} Performance Evaluation",
    author = "Elangovan, Aparna  and
      He, Jiayuan  and
      Verspoor, Karin",
    booktitle = "ACL",
    month = apr,
    year = "2021"
}

@inproceedings{he2021efficient,
    title = "Efficient Nearest Neighbor Language Models",
    author = "He, Junxian  and
      Neubig, Graham  and
      Berg-Kirkpatrick, Taylor",
    booktitle = "Proc. of EMNLP",
    year = "2021",
}

@inproceedings{feldman2020does,
  author    = {Vitaly Feldman},
  title     = {Does learning require memorization? a short tale about a long tail},
  booktitle = { {ACM} {SIGACT}, 2020},
  pages     = {954--959},
  publisher = {{ACM}},
  year      = {2020},
}

@inproceedings{koh2017understanding,
  title={Understanding black-box predictions via influence functions},
  author={Koh, Pang Wei and Liang, Percy},
  booktitle={International Conference on Machine Learning},
  year={2017},
}

@inproceedings{DBLP:conf/iclr/KhandelwalLJZL20,
  author    = {Urvashi Khandelwal and
               Omer Levy and
               Dan Jurafsky and
               Luke Zettlemoyer and
               Mike Lewis},
  title     = {Generalization through Memorization: Nearest Neighbor Language Models},
  booktitle = {{ICLR} 2020},
  publisher = {OpenReview.net},
  year      = {2020}
}

@article{meng2021gnnlm,
  author    = {Yuxian Meng and
               Shi Zong and
               Xiaoya Li and
               Xiaofei Sun and
               Tianwei Zhang and
               Fei Wu and
               Jiwei Li},
  title     = {{GNN-LM:} Language Modeling based on Global Contexts via {GNN}},
  journal   = {ICLR 2022},
  year      = {2022}
}

@misc{alon2022neurosymbolic,
      title={Neuro-Symbolic Language Modeling with Automaton-augmented Retrieval}, 
      author={Uri Alon and Frank F. Xu and Junxian He and Sudipta Sengupta and Dan Roth and Graham Neubig},
      year={2022},
}

@inproceedings{schick2020exploiting,
  author    = {Timo Schick and
               Hinrich Sch{\"{u}}tze},
  title     = {Exploiting Cloze-Questions for Few-Shot Text Classification and Natural
               Language Inference},
  booktitle = {Proceedings of  {EACL} 2021},
  year      = {2021},
}

@inproceedings{shin2020eliciting,
  author    = {Taylor Shin and
               Yasaman Razeghi and
               Robert L. Logan IV and
               Eric Wallace and
               Sameer Singh},
  title     = {AutoPrompt: Eliciting Knowledge from Language Models with Automatically
               Generated Prompts},
  booktitle = {Proceedings of {EMNLP} 2020},
  year      = {2020},
}

@article{p_tuning,
  author    = {Xiao Liu and
               Yanan Zheng and
               Zhengxiao Du and
               Ming Ding and
               Yujie Qian and
               Zhilin Yang and
               Jie Tang},
  title     = {{GPT} Understands, Too},
  journal   = {CoRR},
  volume    = {abs/2103.10385},
  year      = {2021}
}

@inproceedings{gao2020making,
  author    = {Tianyu Gao and
               Adam Fisch and
               Danqi Chen},
  title     = {Making Pre-trained Language Models Better Few-shot Learners},
  booktitle = {Proceedings of ACL},
  year      = {2021},

}

@inproceedings{schick2020automatically,
    title = "Automatically Identifying Words That Can Serve as Labels for Few-Shot Text Classification",
    author = {Schick, Timo  and
      Schmid, Helmut  and
      Sch{\"u}tze, Hinrich},
    booktitle = "Proceedings of COLING",
    month = dec,
    year = "2020",
}

@inproceedings{DBLP:conf/iclr/000100LDC23,
  author       = {Ningyu Zhang and
                  Lei Li and
                  Xiang Chen and
                  Xiaozhuan Liang and
                  Shumin Deng and
                  Huajun Chen},
  title        = {Multimodal Analogical Reasoning over Knowledge Graphs},
  booktitle    = {The Eleventh International Conference on Learning Representations,
                  {ICLR} 2023, Kigali, Rwanda, May 1-5, 2023},
  publisher    = {OpenReview.net},
  year         = {2023}
}

@inproceedings{chen21knowprompt,
  author       = {Xiang Chen and
                  Ningyu Zhang and
                  Xin Xie and
                  Shumin Deng and
                  Yunzhi Yao and
                  Chuanqi Tan and
                  Fei Huang and
                  Luo Si and
                  Huajun Chen},
  title        = {KnowPrompt: Knowledge-aware Prompt-tuning with Synergistic Optimization
                  for Relation Extraction},
  booktitle    = {{WWW} '22: The {ACM} Web Conference 2022, Virtual Event, Lyon, France,
                  April 25 - 29, 2022},
  pages        = {2778--2788},
  publisher    = {{ACM}},
  year         = {2022},
 
}

@article{DBLP:journals/corr/abs-2203-00902,
  author    = {Sen Yang and
               Yunchen Zhang and
               Leyang Cui and
               Yue Zhang},
  title     = {Do Prompts Solve {NLP} Tasks Using Natural Language?},
  journal   = {CoRR},
  volume    = {abs/2203.00902},
  year      = {2022}
}

@article{DBLP:journals/corr/abs-2104-08786,
  author    = {Yao Lu and
               Max Bartolo and
               Alastair Moore and
               Sebastian Riedel and
               Pontus Stenetorp},
  title     = {Fantastically Ordered Prompts and Where to Find Them: Overcoming Few-Shot
               Prompt Order Sensitivity},
  journal   = {CoRR},
  volume    = {abs/2104.08786},
  year      = {2021}
}

@inproceedings{meng2020text,
  title={Text Classification Using Label Names Only: A Language Model Self-Training Approach},
  author={Meng, Yu and Zhang, Yunyi and Huang, Jiaxin and Xiong, Chenyan and Ji, Heng and Zhang, Chao and Han, Jiawei},
  booktitle={Proceedings of EMNLP},
  year={2020}
}

@article{DBLP:journals/tbd/JohnsonDJ21,
  author    = {Jeff Johnson and
               Matthijs Douze and
               Herv{\'{e}} J{\'{e}}gou},
  title     = {Billion-Scale Similarity Search with GPUs},
  year      = {2021}
}

@article{memory,
  author    = {Xiaosen Zheng and
               Jing Jiang},
  title     = {An Empirical Study of Memorization in {NLP}},
  journal   = {CoRR},
  volume    = {abs/2203.12171},
  year      = {2022}
}

@inproceedings{DBLP:conf/cvpr/DengDSLL009,
  author    = {Jia Deng and
               Wei Dong and
               Richard Socher and
               Li{-}Jia Li and
               Kai Li and
               Li Fei{-}Fei},
  title     = {ImageNet: {A} large-scale hierarchical image database},
  booktitle = {{(CVPR} 2009)},
  publisher = {{IEEE} Computer Society},
  year      = {2009}
}

@inproceedings{DBLP:conf/cvpr/LiFP04,
  author    = {Li Fei{-}Fei and
               Rob Fergus and
               Pietro Perona},
  title     = {Learning Generative Visual Models from Few Training Examples: An Incremental
               Bayesian Approach Tested on 101 Object Categories},
  booktitle = {
               {CVPR} Workshops 2004},
  pages     = {178},
  publisher = {{IEEE} Computer Society},
  year      = {2004}
}

@inproceedings{DBLP:conf/cvpr/CimpoiMKMV14,
  author    = {Mircea Cimpoi and
               Subhransu Maji and
               Iasonas Kokkinos and
               Sammy Mohamed and
               Andrea Vedaldi},
  title     = {Describing Textures in the Wild},
  booktitle = {
               {CVPR} 2014},
  pages     = {3606--3613},
  publisher = {{IEEE} Computer Society},
  year      = {2014}
}

@article{DBLP:journals/corr/MajiRKBV13,
  author    = {Subhransu Maji and
               Esa Rahtu and
               Juho Kannala and
               Matthew B. Blaschko and
               Andrea Vedaldi},
  title     = {Fine-Grained Visual Classification of Aircraft},
  journal   = {CoRR},
  volume    = {abs/1306.5151},
  year      = {2013}
}

@inproceedings{DBLP:conf/icvgip/NilsbackZ08,
  author    = {Maria{-}Elena Nilsback and
               Andrew Zisserman},
  title     = {Automated Flower Classification over a Large Number of Classes},
  booktitle = { {ICVGIP} 2008},
  pages     = {722--729},
  publisher = {{IEEE} Computer Society},
  year      = {2008},
}

@inproceedings{DBLP:conf/eccv/BossardGG14,
  author    = {Lukas Bossard and
               Matthieu Guillaumin and
               Luc Van Gool},
  title     = {Food-101 - Mining Discriminative Components with Random Forests},
  booktitle = {{ECCV} 2014,
               Switzerland, September 6-12, 2014, Proceedings, Part {VI}},
  year      = {2014},
}

@inproceedings{DBLP:conf/cvpr/ParkhiVZJ12,
  author    = {Omkar M. Parkhi and
               Andrea Vedaldi and
               Andrew Zisserman and
               C. V. Jawahar},
  title     = {Cats and dogs},
  booktitle = {CVPR, 2012},
  pages     = {3498--3505},
  publisher = {{IEEE} Computer Society},
  year      = {2012}
}

@inproceedings{DBLP:conf/iccvw/Krause0DF13,
  author    = {Jonathan Krause and
               Michael Stark and
               Jia Deng and
               Li Fei{-}Fei},
  title     = {3D Object Representations for Fine-Grained Categorization},
  booktitle = {
               {ICCV} Workshops 2013},
  year      = {2013}
}

@article{DBLP:journals/corr/abs-1212-0402,
  author    = {Khurram Soomro and
               Amir Roshan Zamir and
               Mubarak Shah},
  title     = {{UCF101:} {A} Dataset of 101 Human Actions Classes From Videos in
               The Wild},
  journal   = {CoRR},
  volume    = {abs/1212.0402},
  year      = {2012}
}

@article{DBLP:journals/ijcv/ZhouYLL22,
  author    = {Kaiyang Zhou and
               Jingkang Yang and
               Chen Change Loy and
               Ziwei Liu},
  title     = {Learning to Prompt for Vision-Language Models},
  journal   = {Int. J. Comput. Vis.},
  volume    = {130},
  number    = {9},
  pages     = {2337--2348},
  year      = {2022}
}

@inproceedings{simvlm,
  author    = {Zirui Wang and
               Jiahui Yu and
               Adams Wei Yu and
               Zihang Dai and
               Yulia Tsvetkov and
               Yuan Cao},
  title     = {SimVLM: Simple Visual Language Model Pretraining with Weak Supervision},
  booktitle = { {ICLR}
               2022},
  publisher = {OpenReview.net},
  year      = {2022}
}

@inproceedings{DBLP:conf/acl/Jin0SC022,
  author    = {Woojeong Jin and
               Yu Cheng and
               Yelong Shen and
               Weizhu Chen and
               Xiang Ren},
  title     = {A Good Prompt Is Worth Millions of Parameters: Low-resource Prompt-based
               Learning for Vision-Language Models},
  booktitle = {{ACL} 2022},
  year      = {2022},
}

@article{cpt,
  author    = {Yuan Yao and
               Ao Zhang and
               Zhengyan Zhang and
               Zhiyuan Liu and
               Tat{-}Seng Chua and
               Maosong Sun},
  title     = {{CPT:} Colorful Prompt Tuning for Pre-trained Vision-Language Models},
  journal   = {CoRR},
  volume    = {abs/2109.11797},
  year      = {2021},
}

@inproceedings{CoCoOp,
  author    = {Kaiyang Zhou and
               Jingkang Yang and
               Chen Change Loy and
               Ziwei Liu},
  title     = {Conditional Prompt Learning for Vision-Language Models},
  booktitle = {
               {CVPR} 2022},
  pages     = {16795--16804},
  publisher = {{IEEE}},
  year      = {2022},
}

@article{DBLP:journals/corr/abs-2302-13971,
  author       = {Hugo Touvron and
                  Thibaut Lavril and
                  Gautier Izacard and
                  Xavier Martinet and
                  Marie{-}Anne Lachaux and
                  Timoth{\'{e}}e Lacroix and
                  Baptiste Rozi{\`{e}}re and
                  Naman Goyal and
                  Eric Hambro and
                  Faisal Azhar and
                  Aur{\'{e}}lien Rodriguez and
                  Armand Joulin and
                  Edouard Grave and
                  Guillaume Lample},
  title        = {LLaMA: Open and Efficient Foundation Language Models},
  journal      = {CoRR},
  volume       = {abs/2302.13971},
  year         = {2023},
  url          = {https://doi.org/10.48550/arXiv.2302.13971},
  doi          = {10.48550/arXiv.2302.13971},
  eprinttype    = {arXiv},
  eprint       = {2302.13971},
  bibsource    = {dblp computer science bibliography, https://dblp.org}
}

@article{DBLP:journals/emnlp/qinlibo,
  author       = {Libo Qin and
                  Qiguang Chen and
                  Fuxuan Wei and
                  Shijue Huang and
                  Wanxiang Che},
  title        = {Cross-lingual Prompting: Improving Zero-shot Chain-of-Thought Reasoning
                  across Languages},
  journal      = {EMNLP },
  year         = {2023},

}

@article{DBLP:journals/taslp/QinXWZC23,
  author       = {Libo Qin and
                  Xiao Xu and
                  Lehan Wang and
                  Yue Zhang and
                  Wanxiang Che},
  title        = {Modularized Pre-Training for End-to-End Task-Oriented Dialogue},
  journal      = {{IEEE} {ACM} Trans. Audio Speech Lang. Process.},
  volume       = {31},
  pages        = {1601--1610},
  year         = {2023},
}

@article{rag_survey,
  author       = {Yunfan Gao and
                  Yun Xiong and
                  Xinyu Gao and
                et al},
  title        = {Retrieval-Augmented Generation for Large Language Models: {A} Survey},
  journal      = {CoRR},
  volume       = {abs/2312.10997},
  year         = {2023},
  bibsource    = {dblp computer science bibliography, https://dblp.org}
}

\end{document}